\documentclass[10pt,twocolumn,letterpaper]{article}

\usepackage[pagenumbers]{cvpr} 

\definecolor{cvprblue}{rgb}{0.21,0.49,0.74}
\usepackage[pagebackref,breaklinks,colorlinks,allcolors=cvprblue]{hyperref}

\usepackage{soul}

\usepackage{colortbl}

\usepackage{makecell}
\usepackage{multirow}

\usepackage{wrapfig}
\usepackage{cutwin}
\usepackage{bbding}

\def\paperID{5765} 
\def\confName{CVPR}
\def\confYear{2026}

\title{MonkeyOCR: Document Parsing with a Structure-Recognition-Relation Triplet Paradigm}

\author{
Zhang Li\textsuperscript{1}, Yuliang Liu\textsuperscript{1,\textdagger}, Qiang Liu\textsuperscript{2}, Zhiyin Ma\textsuperscript{1}, Ziyang Zhang\textsuperscript{1}, \\ 
Shuo Zhang\textsuperscript{1},
Biao Yang\textsuperscript{1},
Zidun Guo\textsuperscript{1}, 
Jiarui Zhang\textsuperscript{2},
Xinyu Wang\textsuperscript{1},
Xiang Bai\textsuperscript{1} \\ 
\textsuperscript{1}Huazhong University of Science and Technology, \textsuperscript{2}Kingsoft Office\\
\textdagger~Project lead
}

\begin{document}

\twocolumn[{
\maketitle
\vspace{-1cm}
\begin{center}
\includegraphics[width=0.90\linewidth]{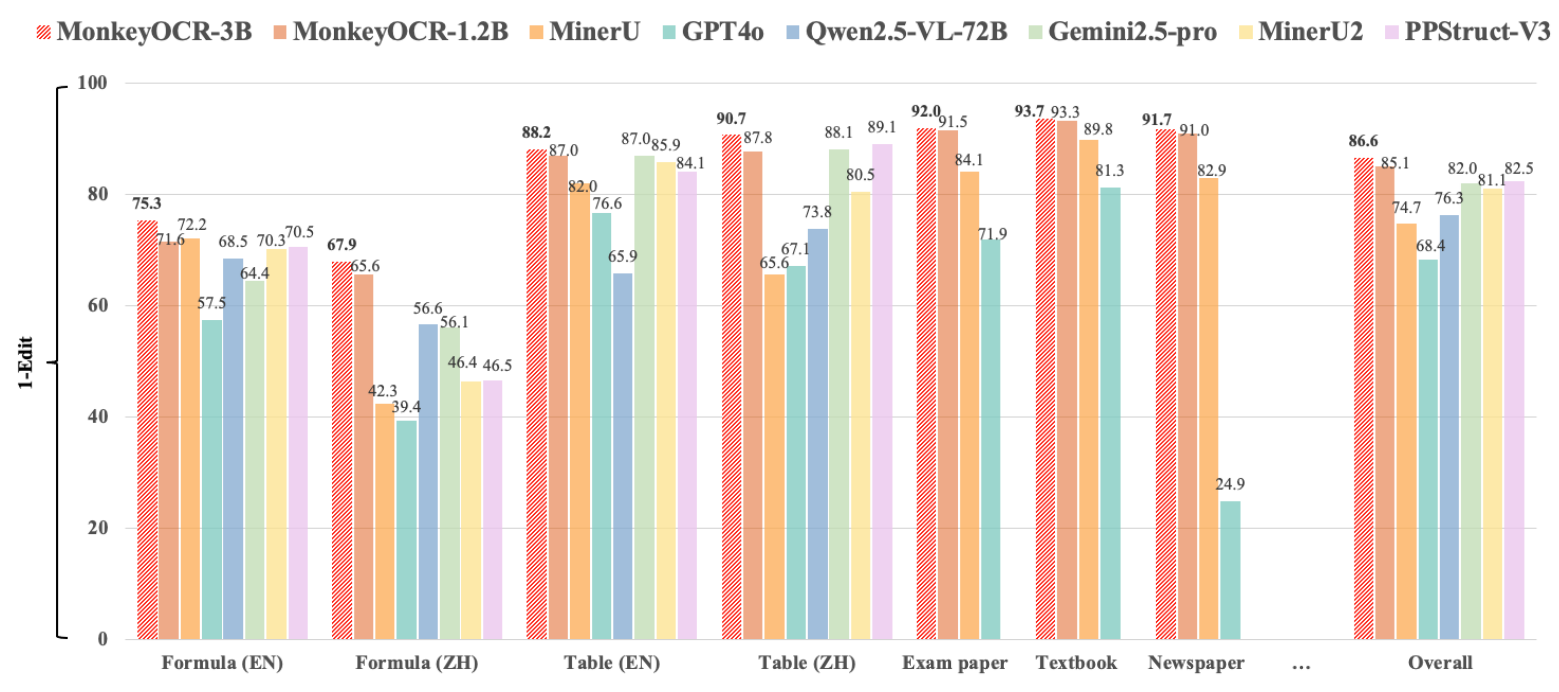}
\end{center}
\vspace{-0.6cm}
\captionsetup{type=figure}
\captionof{figure}{Performance comparison of MonkeyOCR and other SOTA models on OmniDocBench~\cite{omnidocbench}.}
}\label{fig:his}
\vspace{0.3cm}
]

\begin{abstract}
We introduce MonkeyOCR, a document parsing model that advances the state of the art by leveraging a Structure–Recognition–Relation (SRR) triplet paradigm. This design simplifies what would otherwise be a complex multi-tool pipeline and avoids the inefficiencies of processing full pages with giant end-to-end models. In SRR, document parsing is abstracted into three fundamental questions — ``Where is it?'' (structure), ``What is it?'' (recognition), and ``How is it organized?'' (relation) — corresponding to structure detection, content recognition, and relation prediction. To support this paradigm, we present MonkeyDoc, a comprehensive dataset with 4.5 million bilingual instances spanning over ten document types, which addresses the limitations of existing datasets that often focus on a single task, language, or document type. Leveraging the SRR paradigm and MonkeyDoc, we trained a 3B-parameter document foundation model. We further identify parameter redundancy in this model and propose contiguous parameter degradation (CPD), enabling the construction of models from 0.6B to 1.2B parameters that run faster with  acceptable performance drop. MonkeyOCR achieves state-of-the-art performance, surpassing previous open-source and closed-source methods, including Gemini 2.5-Pro. Additionally, the model can be efficiently deployed for inference on a single RTX 3090 GPU. Code and models will be released at \url{https://github.com/Yuliang-Liu/MonkeyOCR}.

\end{abstract}

\begin{figure*}[ht]
\centering
\includegraphics[width=0.9\linewidth]{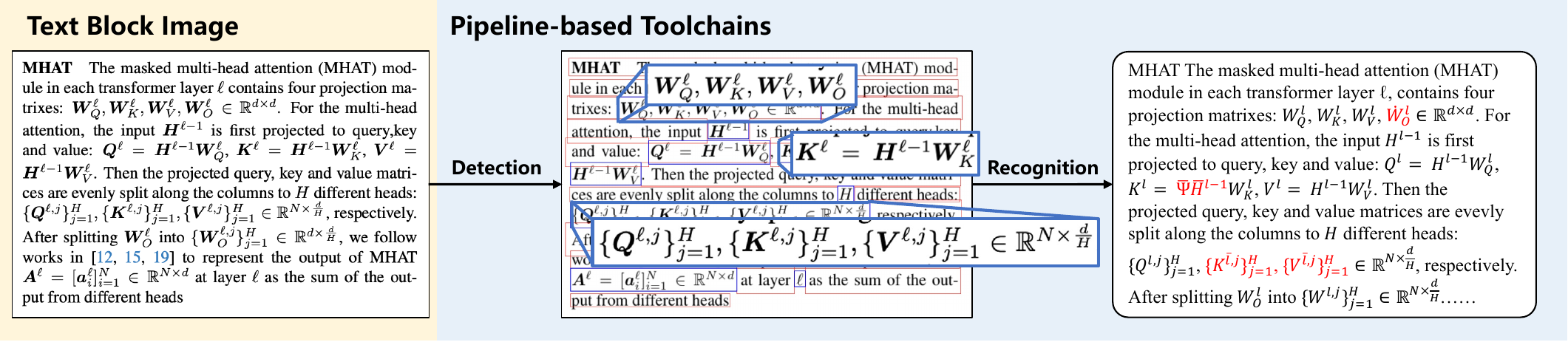}
\caption{A typical example of cumulative error in pipeline-based toolchains.}
\label{fig:intro}
\end{figure*}

\section{Introduction}
Document parsing aims to systematically convert the complex multimodal content (including texts, tables, images, formulas, etc.) within various documents (such as scanned images and PDF files) into structured information. 
This technology has a wide range of applications, including smart office solutions, intelligent education systems, digitization of medical records and judicial documents, and the preservation of historical documents.

Recent mainstream document parsing methods can be broadly categorized into pipeline-based toolchains and end-to-end models. Pipeline-based toolchains, such as MinerU~\cite{mineru} and PPStruct-V3~\cite{ppstruct}, break down the complex parsing task into fine-grained sub-tasks, with each sub-task handled by specialized models. This modular approach allows each component to be optimized independently. In contrast, giant end-to-end models~\cite{qwen2.5-vl} directly inputting entire document page images into a unified model for holistic content recognition, resulting in a streamlined workflow.

Although existing document parsing approaches have achieved remarkable progress, they still suffer from inherent limitations. First, pipeline-based methods suffer from cumulative error caused by the combination of multiple fine-grained tools.
These methods typically follow a sequential process: layout detection, text and formula detection, instance-level recognition, element merging, and finally reading order reconstruction.
As shown in Fig.~\ref{fig:intro}, imprecise formula detection results in partial cropping of characters from the previous line, leading to erroneous recognition outcomes, such as the inclusion of extraneous superscripts. 
Second, end-to-end models often face limitations in both performance and efficiency when dealing with lengthy contexts. Document pages are typically of high resolution and contain densely packed information, leading to extremely long input and output sequences. Prior studies~\cite{longtext1,longtext2} have shown that large multimodal models are more prone to hallucinations when processing long contexts. Moreover, the quadratic computational cost of attention makes the processing of long sequences increasingly expensive.

To achieve a balance between accuracy and speed, we propose MonkeyOCR, which adopts a ``Structure-Recognition-Relation" (SRR) triplet paradigm for document parsing.  Specifically, we first perform structure detection to identify different regions within the document, such as text blocks, tables, formulas, images. Then, to handle multiple recognition tasks with a unified model, we trained a large multimodal model to perform block-level parallel end-to-end recognition of these detected regions. Finally, the coordinate and category of these regions is fed into a block-level reading order prediction model to determine the reading order. Based on the reading order, the results of content recognition are then combined to produce the final document content.

\begin{figure*}[ht]
\centering
\includegraphics[width=0.9\linewidth]{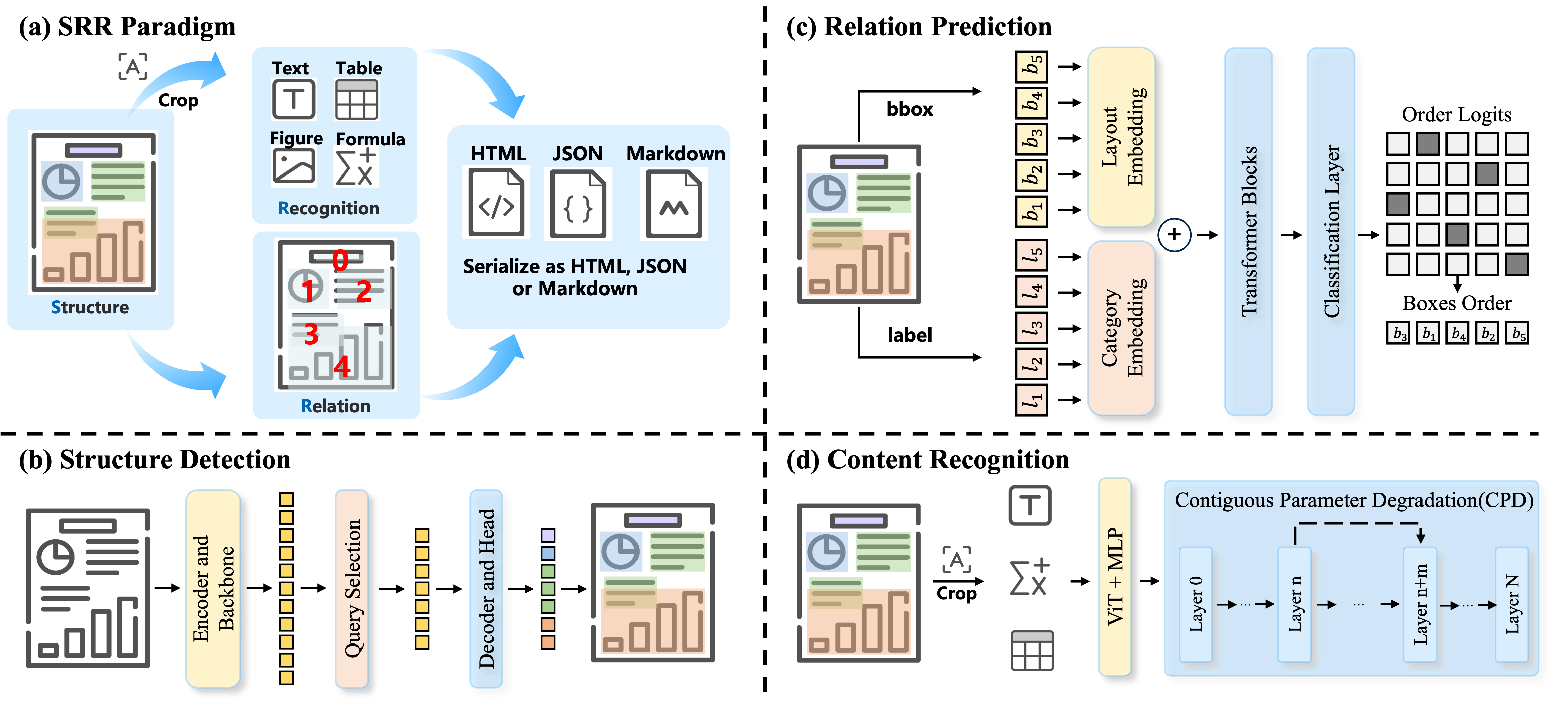}
\vspace{-10pt}
\caption{The overall pipeline of MonkeyOCR.}
\label{fig:method_pipeline}
\vspace{-10pt}
\end{figure*}

Existing document parsing datasets often focus on a single task (e.g., table recognition or layout analysis), a specific document type (e.g., academic papers), or a single language, which limits the development of robust document parsing models.
To address these limitations, we developed a comprehensive data generation pipeline that integrates meticulous manual annotation, programmatic data synthesis, and advanced model-driven automatic labeling, resulting in MonkeyDoc dataset. The dataset contains 4.5 million bilingual instances, covering five document parsing tasks, involving more than ten document types.

Building on the SRR paradigm and the MonkeyDoc dataset, we trained a 3B-parameter document foundation model. We further observed parameter redundancy in this model and, based on experimental findings, empirically propose contiguous parameter degradation (CPD), which continuously skips sequences of intermediate layers to construct lightweight models (0.6B–1.2B parameters) that achieve comparable performance with faster inference. Moreover, our results show that parameter reduction has minimal impact on simple tasks, such as text recognition, but substantially affects more complex tasks, such as table recognition.
Compared to pipeline-based methods, MonkeyOCR reduces the cumulative errors caused by multi-tool combination, outperforming the best pipeline method PPStruct-V3 by 4.1\%. It also surpasses giant end-to-end model Qwen2.5-VL-72B and the best closed-source model Gemini 2.5-Pro, achieving 2.09× faster inference than the end-to-end baseline. Furthermore, MonkeyOCR-1.2B offers a 34\% speed improvement over the 3B model with only a 1.5\% performance drop.
We summarize the contributions as follows:

\begin{itemize}
    \item We propose the SRR paradigm, decomposing document parsing into structure detection, content recognition, and relation prediction. This design reduces cumulative errors caused by multi-tool integration and avoids the inefficiencies of processing full pages with end-to-end models.

    \item We present MonkeyDoc, a comprehensive dataset with 4.5 million bilingual instances covering five document parsing tasks across more than ten document types, addressing the limitations of existing datasets that often focus on a single task, language, or document type.

    \item We propose a 3B document foundation model based on the SRR paradigm and MonkeyDoc, and introduce contiguous parameter degradation (CPD) to prune redundant layers. CPD enables the construction of lightweight models (0.6B–1.2B parameters) that achieve comparable performance while offering faster inference.
    
\end{itemize}

\section{Related Work}

\subsection{Pipeline-based Toolchains}
Pipeline-based approaches~\cite{pix2text,docling,marker,mineru} decompose the task of document parsing into multiple sequential steps, such as layout analysis~\cite{rtdetr,layoutlmv3,yolov10}, reading order prediction~\cite{layoutreader}, optical character recognition (OCR)~\cite{easyocr,ppocr}, formula recognition~\cite{unimernet}, formula detection, and table recognition~\cite{tabformer,tablemaster}. Each step employs a specialized model to handle a specific sub-task.
Docling~\cite{docling} implements a linear processing pipeline to parse PDF files, extract page content (including OCR, layout, and table information), and generate structured documents in JSON or Markdown format.
Marker~\cite{marker} relies on the Surya OCR toolkit~\cite{paruchuri2025surya} for OCR, layout, reading order, and table recognition, with optional LLM modules for cross-page table merging and inline math parsing.
Similarly, MinerU and PPStruct-V3 adopt a pipeline of multiple models for layout detection~\cite{doclayout}, formula detection~\cite{yolov8_ultralytics} and recognition~\cite{unimernet}, OCR~\cite{ppocr}, table recognition~\cite{ye2021pingan, docgenome}, and reading order prediction~\cite{layoutreader, Gu_2022_CVPR}, and integrate these outputs to complete document parsing.

\begin{table*}[t]
\centering
\resizebox{0.9\textwidth}{!}{
\begin{tabular}{@{}c|c|ccccc|cc}
\toprule
\multirow{3}{*}{\textbf{Dataset}} & \multirow{3}{*}{\textbf{\makecell{Document\\ type}}} & \multicolumn{5}{c|}{\textbf{Supporting Tasks}} & \multicolumn{2}{c}{\textbf{Language}} \\
& & \makecell{Layout \\ Detection} & \makecell{Reading Order \\ Prediction} & \makecell{Formula \\ Recognition} & \makecell{Table \\ Recognition} & \makecell{Text \\ Recognition} & EN & ZH \\ \midrule

M\textsuperscript{6}Doc~\cite{m6doc}, DocLayNet~\cite{doclaynet} & 6, 2 & \Checkmark & & & & & \Checkmark & \Checkmark \\
CDLA~\cite{cdla} & 1 & \Checkmark & & & & & & \Checkmark \\
D4LA~\cite{d4la} & 5 & \Checkmark &  & & & & \Checkmark & \\ 
Unimernet~\cite{unimernet}, TexTeller~\cite{texteller} & - & & & \Checkmark & & & &  \\ 
FinTabNet~\cite{fintabnet}, PubTabNet~\cite{pubtabnet} & - & & & & \Checkmark & & &  \\ 
DocGenome~\cite{docgenome} & 1 & \Checkmark & \Checkmark & \Checkmark & \Checkmark & \Checkmark & \Checkmark &  \\
\rowcolor{gray!10}
\textbf{MonkeyDoc} & \textbf{$>$10} & \Checkmark & \Checkmark & \Checkmark & \Checkmark & \Checkmark & \Checkmark & \Checkmark \\ \bottomrule
\end{tabular}
}
\caption{Comparison of MonkeyDoc with other document parsing datasets.}
\label{tab:MonkeyDoc}
\vspace{-2pt}
\end{table*}

\subsection{End-to-end Models}
End-to-end models streamline document parsing by directly processing full-page images into structured outputs, eliminating the need for multiple task-specific models.
Donut~\cite{donut} introduces the first OCR-free framework for document understanding, while Nougat~\cite{nougat} adopts a similar architecture to convert document images into Markdown.
The SPTS series~\cite{spts,sptsv2} presents a unified framework for text detection and recognition, and OmniParser~\cite{omniparser} extends it to text spotting, key information extraction, and table recognition, enabling robust parsing across both natural scenes and documents.
Recently, large multimodal models~\cite{got,monkey,textmonkey,internvl,qwen2.5-vl} trained on massive datasets have achieved remarkable progress in end-to-end document understanding.
Building on Qwen2.5-VL~\cite{qwen2.5-vl}, models such as olmOCR~\cite{olmocr}, OCRFlux~\cite{ocrflux}, and Nanonets-OCR~\cite{Nanonets-OCR-S} are further trained on large corpora of PDF pages, enhancing the document parsing capabilities of general-purpose multimodal models. DeepSeekOCR-3B~\cite{deepseekocr} explores visual token compression techniques for efficient document parsing.


\subsection{Document Parsing Dataset}
Document parsing encompasses fine-grained tasks such as layout detection, reading order prediction, and content recognition. A variety of datasets have been developed to support these tasks, each focusing on specific document types, languages or sub-tasks.
For layout detection, datasets such as M\textsuperscript{6}Doc~\cite{m6doc}, CDLA~\cite{cdla}, D4LA~\cite{d4la}, and DocLayNet~\cite{doclaynet} provide annotations of the positions of structural elements (e.g., text blocks, tables) across diverse document types.
For content recognition, table recognition datasets such as FinTabNet~\cite{fintabnet} and PubTabNet~\cite{pubtabnet} provide large-scale structural and content annotations, while Unimer-1M~\cite{unimernet} and HME-100k~\cite{hme-100k} focus on mathematical expression recognition in printed and handwritten forms.
For reading order prediction, 
ReadingBank~\cite{layoutreader} provides word-level reading order annotations for 500k documents.
Additionally, DocGenome~\cite{docgenome} provides annotations for 500K English papers from arXiv, covering major document parsing tasks.

\begin{figure}[ht]
    \centering
    \includegraphics[width=0.9\linewidth]{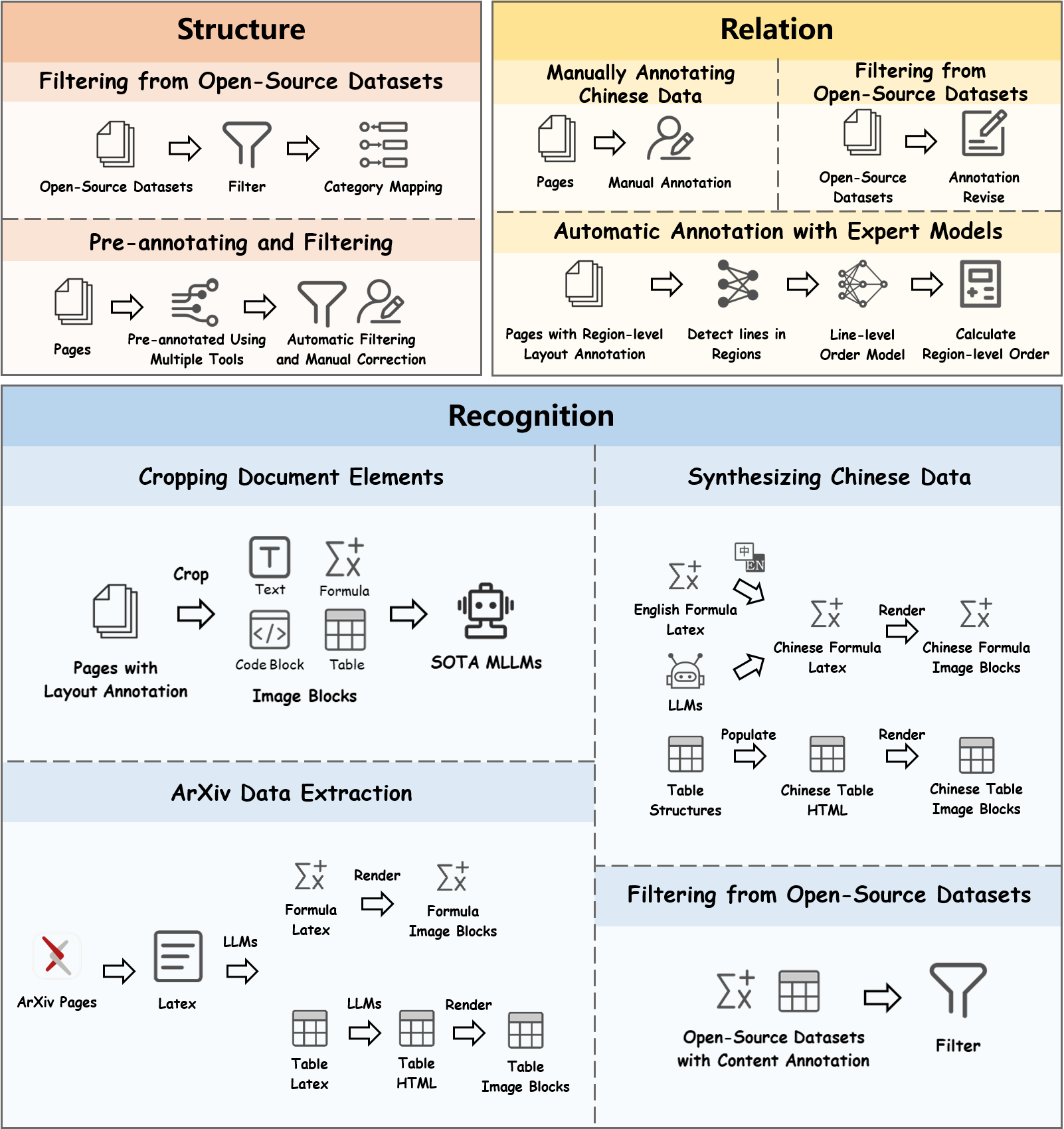}
    \caption{The data generation pipeline of MonkeyDoc.}
    \label{fig:data_gen}
    \vspace{-5pt}
\end{figure}

\begin{figure*}[ht]
\centering
\includegraphics[width=0.95\linewidth]{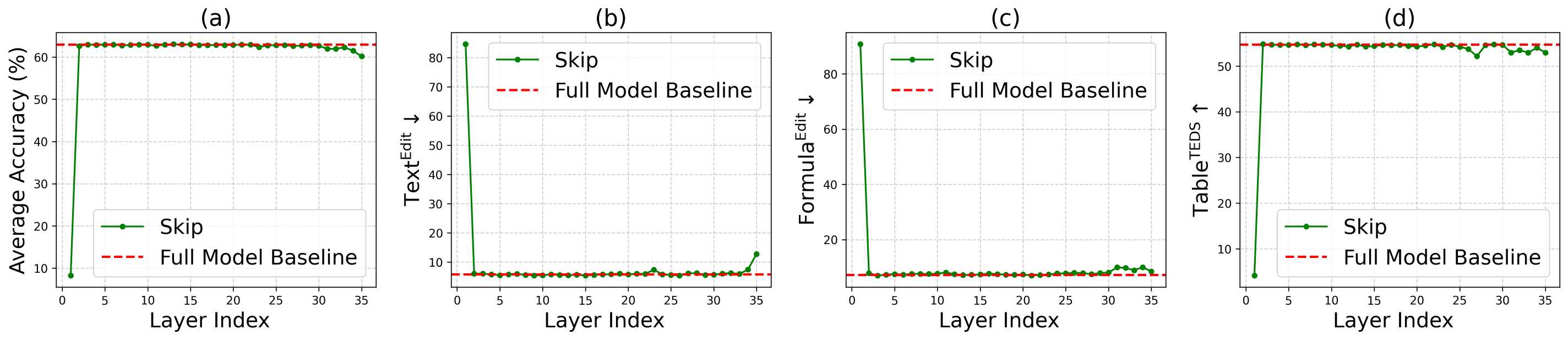}
\caption{Results of MonkeyOCR-3B on text, formula, and table tasks. Green denotes skipping layer N, and red denotes no skipping.}
\label{fig:skip}
\end{figure*}

\section{MonkeyDoc Dataset}

To train MonkeyOCR, we developed a complete data generation pipeline and built a comprehensive document parsing dataset, MonkeyDoc. As shown in Tab.~\ref{tab:MonkeyDoc}, the dataset covers more than ten document types, supports both Chinese and English, and spans the entire document parsing process. The overall data generation pipeline is illustrated in Fig.~\ref{fig:data_gen}. More details can be found in the appendix.

For structure detection, we collected raw annotations from existing open-source datasets~\cite{m6doc, doclaynet, d4la, cdla, docgenome}, removed nested bounding boxes, and selected samples with richer structural elements to enhance data diversity.
Moreover, considering the scarcity of Chinese samples in open-source datasets, we additionally collected over 300k pages Chinese documents, including financial reports, textbooks, academic papers, etc. These pages were pre-annotated using multiple tools~\cite{ppstruct, doclayout}, followed by rule-based automatic filtering and manual correction and validation to ensure annotation accuracy and consistency.

For content recognition, we cropped a total of 2.5 million sub-images of document elements based on the structure detection results, which were then annotated using Qwen2.5-VL-72B~\cite{qwen2.5-vl} and Gemini 2.5 Pro~\cite{gemini25}. To further expand our datasets for table and formula recognition, we employed a rule-based method to filter out erroneous samples from the open-source datasets PubTabNet~\cite{pubtabnet} and UniMER-1M~\cite{unimernet}. Addressing the scarcity of Chinese data, we populated existing table structures with Chinese characters and rendered them into new table images. For formulas, we designed prompts to guide an LLM in generating commonly used Chinese formulas across various domains and also used the LLM to translate existing English formulas into Chinese, which were subsequently rendered. To enhance the diversity and authenticity of our table and formula data, we scraped LaTeX source from ArXiv, used an LLM to filter out irrelevant content, converted the tables to HTML format, and then rendered them into images.

For relation prediction, our goal is to construct a block-level reading order dataset that bridges the gap between existing block-level structure detection and line-level reading order prediction methods. First, we refined the existing open-source dataset~\cite{docgenome} by linking tables and images to their corresponding captions and removing low-quality samples to improve annotation accuracy. Second, to address the scarcity of Chinese documents, we manually annotated a variety of Chinese document types, enriching the dataset's diversity. Finally, to fully utilize the large amount of structure detection data with layout annotations but without reading order information, we adopted an automatic labeling strategy. This strategy first employs an OCR tool~\cite{ppocr} to detect text lines within each block, then uses a line-level reading order model~\cite{layoutreader} to predict their sequence. The block-level reading order is then inferred by calculating the average reading order of all text lines within the block.

\section{MonkeyOCR}
Fig.~\ref{fig:method_pipeline} illustrates the overall pipeline of MonkeyOCR. First, the input image is fed into the structure detection model to obtain bounding boxes and categories for each document element. Based on these bounding boxes, the corresponding text, table, and formula regions are cropped and sent to an LMM for content recognition. Meanwhile, the coordinates and categories of the detected elements are passed to the relation prediction model to determine the reading order of the document elements. Finally, the recognized results are integrated according to the predicted reading order to produce the final structured document. The proposed SRR paradigm simplifies traditional pipeline-based methods into three unified models, thereby mitigating the cumulative errors caused by multiple tool combinations. Moreover, by processing document blocks in parallel, our method avoids the long context lengths associated with full-document processing, leading to substantial gains in speed.  
We also found parameter redundancy in large multimodal models and proposed the contiguous parameter degradation method, which enables faster inference with minimal performance degradation.

\begin{table*}[ht]
\centering
\resizebox{0.9\textwidth}{!}{
\begin{tabular}{c|l|cc|cc|cc|cc|cc|cc}
\toprule
\multirow{2}{*}{\makecell{\textbf{Model}\\ \textbf{Type}}} & \multirow{2}{*}{\textbf{Methods}} 
& \multicolumn{2}{c|}{\textbf{Overall}\textsuperscript{\textbf{Edit}}↓} 
& \multicolumn{2}{c|}{\textbf{Text}\textsuperscript{\textbf{Edit}}↓} 
& \multicolumn{2}{c|}{\textbf{Formula}\textsuperscript{\textbf{Edit}}↓} 
&  \multicolumn{2}{c|}{\textbf{Table}\textsuperscript{\textbf{TEDS}}↑} 
& \multicolumn{2}{c|}{\textbf{Table}\textsuperscript{\textbf{Edit}}↓} 
& \multicolumn{2}{c}{\textbf{Read Order}\textsuperscript{\textbf{Edit}}↓} \\
\cmidrule{3-14}
 &  & \textit{EN} & \textit{ZH} & \textit{EN} & \textit{ZH} & \textit{EN} & \textit{ZH} & \textit{EN} & \textit{ZH} & \textit{EN} & \textit{ZH} & \textit{EN} & \textit{ZH} \\

\midrule
\multirow{8}{*}{\makecell{\textbf{Pipeline}\\ \textbf{Tools}}} & MinerU~\cite{mineru} & 0.150 & 0.357 & 0.061 & 0.215 & 0.278 & 0.577  & 78.6 & 62.1 & 0.180 & 0.344 & {0.079} & 0.292 \\
 & Marker~\cite{marker} & 0.336 & 0.556 & 0.080 & 0.315 & 0.530 & 0.883  & 67.6 & 49.2 & 0.619 & 0.685 & 0.114 & 0.340 \\
 & Mathpix~\cite{mathpix} & 0.191 & 0.365 & 0.105 & 0.384 & 0.306 & {0.454}  & 77.0 & 67.1 & 0.243 & 0.320 & 0.108 & 0.304 \\
 & Docling~\cite{docling} & 0.589 & 0.909 & 0.416 & 0.987 & 0.999 & 1  & 61.3 & 25.0 & 0.627 & 0.810 & 0.313 & 0.837 \\
 & Pix2Text~\cite{pix2text} & 0.320 & 0.528 & 0.138 & 0.356 & 0.276 & 0.611 & 73.6 & 66.2 & 0.584 & 0.645 & 0.281 & 0.499 \\
 & Unstructured~\cite{unstructured} & 0.586 & 0.716 & 0.198 & 0.481 & 0.999 & 1 & 0 & 0.06 & 1 & 0.998 & 0.145 & 0.387 \\
 & OpenParse~\cite{openparse} & 0.646 & 0.814 & 0.681 & 0.974 & 0.996 & 1  & 64.8 & 27.5 & 0.284 & 0.639 & 0.595 & 0.641 \\

 & PPStruct-V3~\cite{ppstruct} & 0.145 & 0.206 & 0.058 & {0.088} & 0.295 & 0.535  & 77.2 & 83.9 & 0.159 & 0.109 & {0.069} & \textbf{0.091} \\
 
\midrule
\multirow{10}{*}{\makecell{\textbf{Expert}\\ \textbf{VLMs}}} & GOT-OCR~\cite{got} & 0.287 & 0.411 & 0.189 & 0.315 & 0.360 & 0.528  & 53.2 & 47.2 & 0.459 & 0.520 & 0.141 & 0.280 \\
 & Nougat~\cite{nougat} & 0.452 & 0.973 & 0.365 & 0.998 & 0.488 & 0.941  & 39.9 & 0 & 0.572 & 1.000 & 0.382 & 0.954 \\
 & Mistral OCR~\cite{mistralocr} & 0.268 & 0.439 & 0.072 & 0.325 & 0.318 & 0.495  & 75.8 & 63.6 & 0.600 & 0.650 & 0.083 & 0.284 \\
 & OLMOCR~\cite{olmocr} & 0.326 & 0.469 & 0.097 & 0.293 & 0.455 & 0.655 & 68.1 & 61.3 & 0.608 & 0.652 & 0.145 & 0.277 \\
 & SmolDocling~\cite{smoldocling} & 0.493 & 0.816 & 0.262 & 0.838 & 0.753 & 0.997  & 44.9 & 16.5 & 0.729 & 0.907 & 0.227 & 0.522 \\

& Dolphin~\cite{dolphin} & 0.206 & 0.306 & 0.107 & 0.197 & 0.447 & 0.580  & 77.3 & 67.2 & 0.180 & 0.285 & 0.091 & 0.162\\
& MinerU2.0~\cite{mineru} & 0.139 & 0.240 & 0.047 & 0.109 & 0.297 & 0.536  & {82.5} & 79.0 & 0.141 & 0.195 & {0.069} & {0.118} \\

& OCRFlux-3B~\cite{ocrflux} & 0.195 & 0.281 & 0.064 & 0.183 & 0.379 & 0.613  & 71.6 & 81.3 & 0.253 & 0.139 & 0.086 & 0.187 \\
& Nanonets-OCR-3B~\cite{Nanonets-OCR-S} &0.283& 0.295 &0.134&0.231&0.518&0.546&76.8&79.4&0.343&0.201&0.135&0.200\\
& DeepSeekOCR-3B-Large~\cite{deepseekocr} & 0.138& 0.208 &0.054&  0.143 &0.277 &0.461&-&-& 0.152&0.104& 0.067&0.123\\
\midrule
\multirow{5}{*}{\makecell{\textbf{General}\\ \textbf{VLMs}}}

 & GPT4o~\cite{gpt4o} & 0.233 & 0.399 & 0.144 & 0.409 & 0.425 & 0.606  & 72.0 & 62.9 & 0.234 & 0.329 & 0.128 & 0.251 \\
 & Qwen2.5-VL-7B~\cite{qwen2.5-vl} & 0.312 & 0.406 & 0.157 & 0.228 & 0.351 & 0.574  & 76.4 & 72.2 & 0.588 & 0.619 & 0.149 & 0.203 \\
 & InternVL3-8B~\cite{internvl} & 0.314 & 0.383 & 0.134 & 0.218 & 0.417 & 0.563  & 66.1 & 73.1 & 0.586 & 0.564 & 0.118 & 0.186 \\
 & Qwen2.5-VL-72B~\cite{qwen2.5-vl} & 0.214 & 0.261 & 0.092 & 0.180 & 0.315 & 0.434  & 82.9 & 83.9 & 0.341 & 0.262 & 0.106 & 0.168 \\
 & Gemini2.5-Pro~\cite{gemini25} & 0.148 & 0.212 & 0.055 & 0.168 & 0.356 & 0.439  & \textbf{85.8} & 86.4 & 0.130 & 0.119 & \textbf{0.049} & 0.121 \\
\midrule

 \multirow{2}{*}{\textbf{Mix}}

 & \textbf{MonkeyOCR-1.2B} & 0.133 & 0.166 & 0.047 & 0.087 & 0.284 & 0.344  & 83.0 & {85.8} & {0.130} & {0.122} & 0.069 & 0.112 \\
 & \textbf{MonkeyOCR-3B} & \textbf{0.118} & \textbf{0.151} & \textbf{0.043} & \textbf{0.077} & \textbf{0.247} & \textbf{0.321}  & {84.2} & \textbf{89.5} & \textbf{0.118} & \textbf{0.093} & {0.066} & 0.113 \\

\bottomrule
\end{tabular}}
\caption{The end-to-end evaluation results of different tasks on OmniDocBench.}
\label{tab:multitasks}
\end{table*}

\subsection{Structure Detection}
We employ DETR-based~\cite{rtdetr} model for structure detection, which consists of a backbone, an encoder, a query selection module, , as well as a decoder and prediction heads.
Given a document image \( I \in \mathbb{R}^{H \times W \times 3} \), the backbone and encoder extracts visual features:
\begin{equation}
F_{0} = \text{Backbone}(I),  
F_{\text{enc}} = \text{Encoder}(F_{0}), 
\end{equation}
where \( F_{\text{enc}} \in \mathbb{R}^{N \times C} \) represents the encoded feature tokens.

To estimate the likelihood of each token corresponding to a foreground region, a lightweight classification head is applied to  $F_{\text{enc}}$:
\begin{equation}
s = \text{SiLU}(\text{BN}(\text{Conv}_{1\times1}(F_{\text{enc}}))),
\end{equation}
where \( s \) represents the foreground probability of each token, indicating its likelihood of belonging to meaningful elements (e.g., text, table, figure).

Based on these scores, the query selection module picks the top-\(K\) most informative features:
\begin{equation}
Q_0 = \text{TopK}(F_{\text{enc}}, s, K),
\end{equation}
effectively filtering background noise and providing high-quality queries for the decoder.

Finally, the decoder refines the selected queries through iterative attention, followed by task-specific heads to produce the final layout predictions.
\begin{equation}
y = \text{Head}(\text{Decoder}(Q_0, F_{\text{0}})),
\end{equation}
where each prediction \( y_i = (b_i, l_i) \) includes a bounding box \( b_i \) and a category label \( l_i \).

\subsection{Content Recognition}
Given the detected bounding boxes \( B = \{b_1, b_2, \dots, b_n\} \), each region is first cropped from the original document image \( I \) to obtain sub-images:
\begin{equation}
I_{\text{crop}}^i = \text{Crop}(I, b_i).
\end{equation}
Each cropped image is preserved at its original resolution and then encoded into a sequence of visual tokens through a vision encoder followed by an MLP projection:
\begin{equation}
T_v^i = \text{MLP}(\text{VisionEnc}(I_{\text{crop}}^i)).
\end{equation}
For each block, a type-specific prompt \( p_{l_i} \) is selected according to its predicted category \( l_i \) (e.g., text, table, or formula). 
Finally, the visual tokens and the corresponding prompt tokens are jointly fed into an LLM to produce the final content predictions:
\begin{equation}
C = \text{LLM}(\{T_v^1, \dots, T_v^n\}, \{p_{l_1}, \dots, p_{l_n}\}),
\end{equation}
where \( C = \{c_1, c_2, \dots, c_n\} \) denotes the textual representations of the detected document elements.

\subsection{Relation Prediction}
We build a block-level category-aware relation prediction model, which consists of embedding layers, transformer blocks and a classification layer. Given a bounding box $b_i = (x_1, y_1, x_2, y_2)$ and its width and height, its positional embedding is computed as
\begin{equation}
P_i = \text{Concat}\big(E(p) \mid p \in \{x_1, y_1, x_2, y_2, w, h\}\big),
\end{equation}
where $E(\cdot)$ denotes separate embedding layers. Different types of document elements typically exhibit distinct semantic priorities for reading order. To explicitly capture this, we propose a category-aware embedding and combine it with positional encoding to construct more discriminative input features:
\begin{equation}
F_i = P_i + E_{\text{c}}(l_i),
\end{equation}
where \(l_i\) denotes the category of the element.

The sequence of features $[F_1, \dots, F_N]$ is processed by transformer layers to model contextual relationships, followed by a classification layer producing the reading order logits:
\begin{equation}
L = \text{Classifier}(\text{Transformer}([F_1, \dots, F_N])),
\end{equation}
where $L$ represents the predicted logits for each element's position. The final reading order is obtained via iterative greedy decoding: each element is first assigned to its highest-scoring position; conflicts are resolved by retaining the element with the highest logit and reassigning others to their next-best positions. This process repeats until a valid permutation is obtained.

\begin{table*}[ht]
  \centering
  \resizebox{1\textwidth}{!}{
    \begin{tabular}{c|l|ccccccccc|c}
    \toprule

  \makecell{\textbf{Model}\\ \textbf{Type}} & \textbf{Models} & \textbf{Book} & \textbf{Slides} & \makecell{\textbf{Financial}\\\textbf{Report}}  & \makecell{\textbf{Textbook}} & \makecell{\textbf{Exam}\\\textbf{Paper}}
  & \textbf{Magazine} & \makecell{\textbf{Academic}\\\textbf{Papers}} & \textbf{Notes}   & \textbf{Newspaper} & \textbf{Overall} \\
     \midrule
     \multirow{3}{*}{\makecell{\textbf{Pipeline}\\ \textbf{Tools}}} & MinerU~\cite{mineru} & \underline{0.055} & 0.124 & {0.033} & {0.102} & {0.159} & \underline{0.072} & \textbf{0.025} & 0.984 & 0.171 & 0.206 \\
     & Marker~\cite{marker} & 0.074 & 0.340 & 0.089 & 0.319 & 0.452 & 0.153 & 0.059 & 0.651 & 0.192 & 0.274 \\
     & Mathpix~\cite{mathpix} & 0.131 & 0.220 & 0.202 & 0.216 & 0.278 & 0.147 & 0.091 & 0.634 & 0.690 & 0.300 \\
    \midrule  
     \multirow{4}{*}{\makecell{\textbf{Expert}\\ \textbf{VLMs}}} & GOT-OCR~\cite{got} & 0.111 & 0.222 & 0.067 & 0.132 & 0.204 & 0.198 & 0.179 & 0.388 & 0.771 & 0.267 \\
     & Nougat~\cite{nougat} & 0.734 & 0.958 & 1.000 & 0.820 & 0.930 & 0.830 & 0.214 & 0.991 & 0.871 & 0.806 \\
     & Dolphin~\cite{dolphin} & 0.091 & 0.131 & 0.057 & 0.146 & 0.231 & 0.121 & 0.074 & 0.363 & 0.307 & 0.177 \\
     & OCRFlux-3B~\cite{ocrflux} & {0.068} & 0.125 & 0.092 & 0.102 & 0.119 & 0.083 & 0.047 & 0.223 & 0.536 & 0.149 \\
     \midrule
     \multirow{3}{*}{\makecell{\textbf{General}\\ \textbf{VLMs}}} 
     & GPT4o~\cite{gpt4o} & 0.157 & 0.163 & 0.348 & 0.187 & 0.281 & 0.173 & 0.146 & 0.607 & 0.751 & 0.316 \\

     & Qwen2.5-VL-7B~\cite{qwen2.5-vl} & 0.148 & \textbf{0.053} & 0.111 & 0.137 & 0.189 & 0.117 & 0.134 & 0.204 & 0.706 & 0.205 \\
     & InternVL3-8B~\cite{internvl} & 0.163 & \underline{0.056} & 0.107 & 0.109 & {0.129} & 0.100 & 0.159 & \underline{0.150} & 0.681 & 0.188 \\
     \midrule
     \multirow{2}{*}{\textbf{Mix}}     
   
   &\textbf{MonkeyOCR-1.2B} & \textbf{0.048} & 0.079 & \textbf{0.014} & \underline{0.067} & \underline{0.085} & \textbf{0.057} & {0.042} & 0.184 & \underline{0.090} & \underline{0.078} \\
   &\textbf{MonkeyOCR-3B} & \textbf{0.048} & 0.068 & \underline{0.015} & \textbf{0.063} & \textbf{0.080} & \textbf{0.057} & \underline{0.040} & \textbf{0.120} & \textbf{0.083} & \textbf{0.067} \\
        
    \bottomrule
    \end{tabular}%
  }
      \vspace{-2pt}
    \caption{The end-to-end text recognition performance on OmniDocBench across 9 PDF page types.}    
    \vspace{-5pt}
  \label{tab:multitypes}%
\end{table*}%

\subsection{Contiguous Parameter Degradation}
In the field of document intelligence, document parsing serves as an upstream task where processing speed is critical. Since models with fewer parameters generally achieve faster inference, we pose the following question: \textit{are all layers in a large multimodal model truly necessary?}
To investigate this, we randomly selected 50 samples from each of the three tasks: text recognition, formula recognition, and table recognition, for experimentation. We systematically pruned the LLM one layer at a time and evaluated the performance on each task. The results, shown in Fig.~\ref{fig:skip}, reveal that removing the first layer of the LLM leads to a notable performance drop, whereas pruning other layers results in relatively minor degradation.
This observation suggests that not all layers in the LLM are indispensable, and that certain parameters may be redundant. However, how to effectively prune the model remains a question. To this end, we conduct a systematic ablation study on different pruning strategies in Sec~\ref{sec:abla}. Based on the results, we empirically propose contiguous parameter degradation, which directly prunes $m$ contiguous middle layers of the LLM followed by fine-tuning. 
We hypothesize that adjacent layers in the LLM carry continuous and tightly coupled information. By performing contiguous pruning primarily on the middle layers, we can maximally preserve the continuity and semantic relationships between neighboring layers, thereby maintaining performance.

\subsection{Implement Details}
MonkeyOCR is initialized from pretrained models~\cite{ppocrvl, qwen2.5-vl,layoutreader} and subsequently undergoes large-scale trained on MonkeyDoc. During the training process, we utilize the AdamW optimizer with a learning rate of 2e-5 and a cosine learning rate schedule. We employ a batch size of 64. Our 3B model was trained on 32 A800 GPUs.

\section{Experiments}
To validate the effectiveness of MonkeyOCR, we conducted a comprehensive comparison with both open-source and closed-source methods on OmniDocBench~\cite{omnidocbench}.

\subsection{Comparison on Different Tasks}

Document parsing encompasses a variety of sub-tasks, including text recognition, formula recognition, table recognition, reading order prediction, and more. To evaluate MonkeyOCR’s performance across these tasks, we compared it with widely-used methods on OmniDocBench~\cite{omnidocbench}, including pipeline tools~\cite{mineru,marker}, expert VLMs~\cite{got,mistralocr}, closed-source general VLMs~\cite{gpt4o}, and open-source general VLMs~\cite{internvl,qwen2.5-vl}. As shown in Tab.~\ref{tab:multitasks}, MonkeyOCR achieves the best overall performance on both Chinese and English document parsing tasks, attaining state-of-the-art results on 7 out of 10 sub-tasks. Specifically, our 3B model surpasses the best pipeline-based method PPStruct-V3, achieving a 13.1\% improvement in formula recognition and outperforming the giant closed-source end-to-end model Gemini2.5-Pro. Moreover, our 1.2B model attains the second-best overall performance, outperforming both DeepSeekOCR-3B and Nanonets-OCR.

\subsection{Comparison Across Document Types}
As shown in Table~\ref{tab:multitypes}, MonkeyOCR achieves the best overall performance.
In particular, the MonkeyOCR-3B model outperforms both large-scale general VLMs and pipeline-based systems, exceeding InternVL3-8B~\cite{internvl} by 12.1\% and MinerU~\cite{mineru} by 13.9\% in overall performance. Remarkably, MonkeyOCR-1.2B model also delivers highly competitive results, ranking second overall with a score of 0.078, only 1.1\% behind the 3B model.
Beyond overall performance, MonkeyOCR demonstrates strong robustness across different document categories, with the 3B model achieving the best results on seven types of documents. These results indicate that our model performs well across different document types.

\subsection{Ablation Study}
\label{sec:abla}

\begin{table}[h]
\centering
\resizebox{\linewidth}{!}{
\begin{tabular}{lccccc}
\toprule
\textbf{Method} & 
\textbf{Text}\textsuperscript{\textbf{Edit}}↓ & 
\textbf{Formula}\textsuperscript{\textbf{Edit}}↓ & 
\textbf{Table}\textsuperscript{\textbf{TEDS}}↑ & 
\textbf{Order}\textsuperscript{\textbf{Edit}}↓ & 
\textbf{Overall}\textsuperscript{\textbf{Edit}}↓ \\
\midrule
Baseline   & 0.349 & 0.525 & 66.2 & 0.341 & 0.461 \\
\rowcolor{gray!3}
+SRR       & 0.071 & 0.333 & 81.0 & 0.157 & 0.175 \\
+MonkeyDoc & \textbf{0.060} & \textbf{0.284} & \textbf{86.9} & \textbf{0.090} & \textbf{0.135} \\
\bottomrule
\end{tabular}}
\caption{Ablation study on the proposed SRR paradigm and the MonkeyDoc dataset.}
\label{tab:abla_srr}
\end{table}

\noindent
\textbf{Ablation study on SRR and MonkeyDoc}. 
We conducted experiments to systematically evaluate the effectiveness of the proposed SRR paradigm and the impact of training on the MonkeyDoc dataset. The baseline is the original Qwen2.5-VL-3B model with end-to-end prediction. As shown in Table~\ref{tab:abla_srr}, SRR significantly enhances the zero-shot document parsing performance of large multimodal models, achieving an average overall improvement of 28.6\%, including a 27.8\% gain in text recognition. This improvement can be attributed to SRR’s ability to mitigate interference from long contexts during inference.
Moreover, after training on the MonkeyDoc dataset, the overall performance increases by 4\%, with a 5.9\% improvement in table recognition and a 6.7\% improvement in reading order, demonstrating the effectiveness of the MonkeyDoc dataset.

\begin{table}[h]
\centering
\resizebox{0.9\linewidth}{!}{
\begin{tabular}{c|
cc|cc|cc}
\toprule
& \multicolumn{2}{c|}{\textbf{Text}\textsuperscript{\textbf{Edit}}↓} 
& \multicolumn{2}{c|}{\textbf{Formula}\textsuperscript{\textbf{Edit}}↓} 
& \multicolumn{2}{c}{\textbf{Overall}\textsuperscript{\textbf{Edit}}↓} \\
\cmidrule(lr){2-7}
& \textit{EN} & \textit{ZH} 
& \textit{EN} & \textit{ZH} 
& \textit{EN} & \textit{ZH} \\ 
\midrule
SRR & 0.043 & 0.077 & 0.247 & 0.321 & 0.118 & 0.151 \\
+TD\&FD & 0.067 & 0.137 & 0.270 & 0.594 & 0.131 & 0.235 \\
\bottomrule
\end{tabular}}
\caption{Impact of SRR on error accumulation. ``TD'' and ``FD'' denote the text and formula detection models.}
\label{tab:abla_cum_error}
\end{table}

\noindent
\textbf{Impact of SRR paradigm on error accumulation}. To evaluate the effect of the proposed SRR paradigm on cumulative errors, we extended MonkeyOCR by incorporating text and formula detection models following the MinerU. Under this setup, text blocks are no longer fed directly into the LMM for recognition. Instead, word-level bounding boxes are first obtained via the detection models, and the content within each box is then recognized by the LMM, with the results subsequently merged. As shown in Table~\ref{tab:abla_cum_error}, introducing these additional detection models leads to a decrease in text recognition by 4.2\% and formula recognition by 14.8\%, demonstrating the effectiveness of SRR in mitigating cumulative errors.

\begin{table}[h]
\centering
\resizebox{0.9\linewidth}{!}{
\begin{tabular}{l|ccccc}
\toprule
\textbf{Model} & \textbf{3090} & \textbf{4090} & \textbf{A6000} & \textbf{H800} & \textbf{Avg.} \\
\midrule
MonkeyOCR-3B   & 0.497 & 1.006 & 0.609 & 0.897 & 0.752 \\
MonkeyOCR-1.2B & 0.677 & 1.436 & 0.825 & 1.101 & 1.010 \\
Baseline-3B    & 0.244 & 0.387 & 0.285 & 0.518 & 0.359 \\
\bottomrule
\end{tabular}}
\caption{Inference speed (pages/s) comparison on different GPUs.}
\label{tab:gpu_performance}
\end{table}

\noindent
\textbf{Analysis of inference speed}. We randomly sampled 300 pages of PDFs from arXiv to analyze the inference speed of the models, with the results summarized in Tab.~\ref{tab:gpu_performance}. Baseline-3B corresponds to Qwen2.5-VL-3B model finetuned on MonkeyDoc, which performs end-to-end prediction directly. The results show that MonkeyOCR-1.2B achieves a 34\% speedup over MonkeyOCR-3B, while incurring only a 1.5\% drop in performance. Furthermore, by enabling parallel block-wise prediction, MonkeyOCR-3B runs 2.09× faster than the fully end-to-end baseline, highlighting the efficiency advantage of the SRR paradigm.

\begin{table}[h]
\centering
\resizebox{0.9\linewidth}{!}{
\begin{tabular}{c|cc|cc|cc}
\toprule
& \multicolumn{2}{c|}{\textbf{Text}\textsuperscript{\textbf{Edit}}↓} 
& \multicolumn{2}{c|}{\textbf{Order}\textsuperscript{\textbf{Edit}}↓} 
& \multicolumn{2}{c}{\textbf{Overall}\textsuperscript{\textbf{Edit}}↓} \\
\cmidrule(lr){2-3} \cmidrule(lr){4-5} \cmidrule(lr){6-7}
& \textit{EN} & \textit{ZH} 
& \textit{EN} & \textit{ZH} 
& \textit{EN} & \textit{ZH} \\ 
\midrule
Baseline & 0.055 & 0.079 & 0.156 & 0.145 & 0.142 & 0.163 \\
+BLRO    & 0.045 & 0.078 & 0.066 & 0.123 & 0.119 & 0.158 \\
+CAE      & 0.043 & 0.077 & 0.066 & 0.113 & 0.118 & 0.151 \\
\bottomrule
\end{tabular}
}
\caption{Ablation study on block-level reading order data (BLRO) and category-aware embedding (CAE).}
\label{tab:abla_reading_order}
\end{table}

\noindent
\textbf{Ablation study on block-level reading order data and category-aware embedding}. The baseline model refers to LayoutReader~\cite{mineru}, which only supports line-level reading order prediction. As shown in Tab.~\ref{tab:abla_reading_order}, training on our constructed block-level reading order data improves the model’s reading order performance for both Chinese and English by 5.6\%. Incorporating category information into the original model further enhances reading order accuracy. Moreover, since the reading order prediction model does not affect the recognition results of other tasks, it only causes a minor impact on other tasks.

\begin{table}[h]
\centering
\resizebox{\linewidth}{!}{
\begin{tabular}{l|cccc}
\toprule
\textbf{Strategy} 
& \textbf{Text}\textsuperscript{\textbf{Edit}}↓ 
& \textbf{Formula}\textsuperscript{\textbf{Edit}}↓ 
& \textbf{Table}\textsuperscript{\textbf{TEDS}}↑ 
& \textbf{Overall}\textsuperscript{\textbf{Edit}}↓ \\
\midrule
Train from scratch     & 0.120 & 0.355 & 65.2 & 0.208 \\
Prune less important layers~\cite{paper4} & 0.085 & 0.336 & 78.0 & 0.170 \\
Prune deeper layers~\cite{paper3}  & 0.101 & 0.385 & 78.1 & 0.188 \\
Prune shallow layers  & 0.091 & 0.363 & 77.9 & 0.180 \\

CPD   & \textbf{0.067} & \textbf{0.314} & \textbf{84.4} & \textbf{0.150} \\
\bottomrule
\end{tabular}}
\caption{Ablation study on pruning strategies.}
\label{tab:abla_pds}
\vspace{-0.2cm}
\end{table}

\noindent
\textbf{Ablation study on pruning strategies}. Previous works~\cite{paper4,paper3} have explored different strategies for pruning LLMs. Shortgpt~\cite{paper4} tends to prune less important layers, while prior work~\cite{paper3} favors pruning deeper layers (11-34). In addition, we further investigate the effects of pruning shallow layers (1-24) as well as training the LLM from scratch.
As shown in Table~\ref{tab:abla_pds}, our proposed CPD outperforms training from scratch by 5.8\% and surpasses the other pruning strategies.

\begin{table}[h]
\centering
\resizebox{\linewidth}{!}{
\begin{tabular}{cc|cccc}
\toprule
\textbf{Layers} & \textbf{Params} & \textbf{Text}\textsuperscript{\textbf{Edit}}↓ & \textbf{Formula}\textsuperscript{\textbf{Edit}}↓ & \textbf{Table}\textsuperscript{\textbf{TEDS}}↑ & \textbf{Overall}\textsuperscript{\textbf{Edit}}↓ \\ 
\midrule
36 & 3.0 & {0.060} & {0.284} & {86.9} & 0.135 \\
12 & 1.2 & 0.067 & 0.314 & 84.4 & 0.150 \\
8 & 0.9 & 0.076 & 0.321 & 78.3 & 0.164 \\
4 & 0.6 & 0.091 & 0.340 & 74.9 & 0.180 \\
\bottomrule
\end{tabular}}
\caption{Ablation study on CPD.}
\label{tab:abla_pdm}
\end{table}

\noindent
\textbf{Ablation study on CPD}. We conduct a ablation study to investigate the impact of degree of parameter degradation on model performance. Specifically, we evaluate model variants retaining only 4, 8, and 12 layers of the original LLM. As shown in Tab.~\ref{tab:abla_pdm}, retaining 12 layers results in only a marginal 1.5\% overall performance drop. However, when only 4 layers are retained, performance on the text recognition decreases by 3.1\%, while the more complex table recognition task suffers a significant 12\% decline. These findings indicate that while smaller models are sufficient for simpler tasks, a larger parameter capacity is crucial for maintaining performance on complex tasks.

\section{Conclusion}
This paper introduces MonkeyOCR, a document parsing model based on the Structure-Recognition-Relation triplet paradigm, which unifies structural detection, content recognition, and relation prediction into a streamlined framework. This design simplifies traditional multi-tool pipelines while avoiding the inefficiencies of directly processing entire pages with LMMs, enabling both high accuracy and efficient deployment. We further identify parameter redundancy in large multimodal models and propose contiguous parameter degradation, which accelerates inference with minimal performance drop. Experiments show that MonkeyOCR achieves state-of-the-art performance on OmniDocBench, surpassing the closed-source commercial model Gemini2.5-Pro, demonstrating its potential as a foundation model for text-centric applications.

{
    \small
    \bibliographystyle{ieeenat_fullname}
    \bibliography{main}
}


\clearpage
\onecolumn
\appendix

\section{MonkeyDoc Dataset}
\begin{figure*}[ht]
\centering
\includegraphics[width=0.92\linewidth]{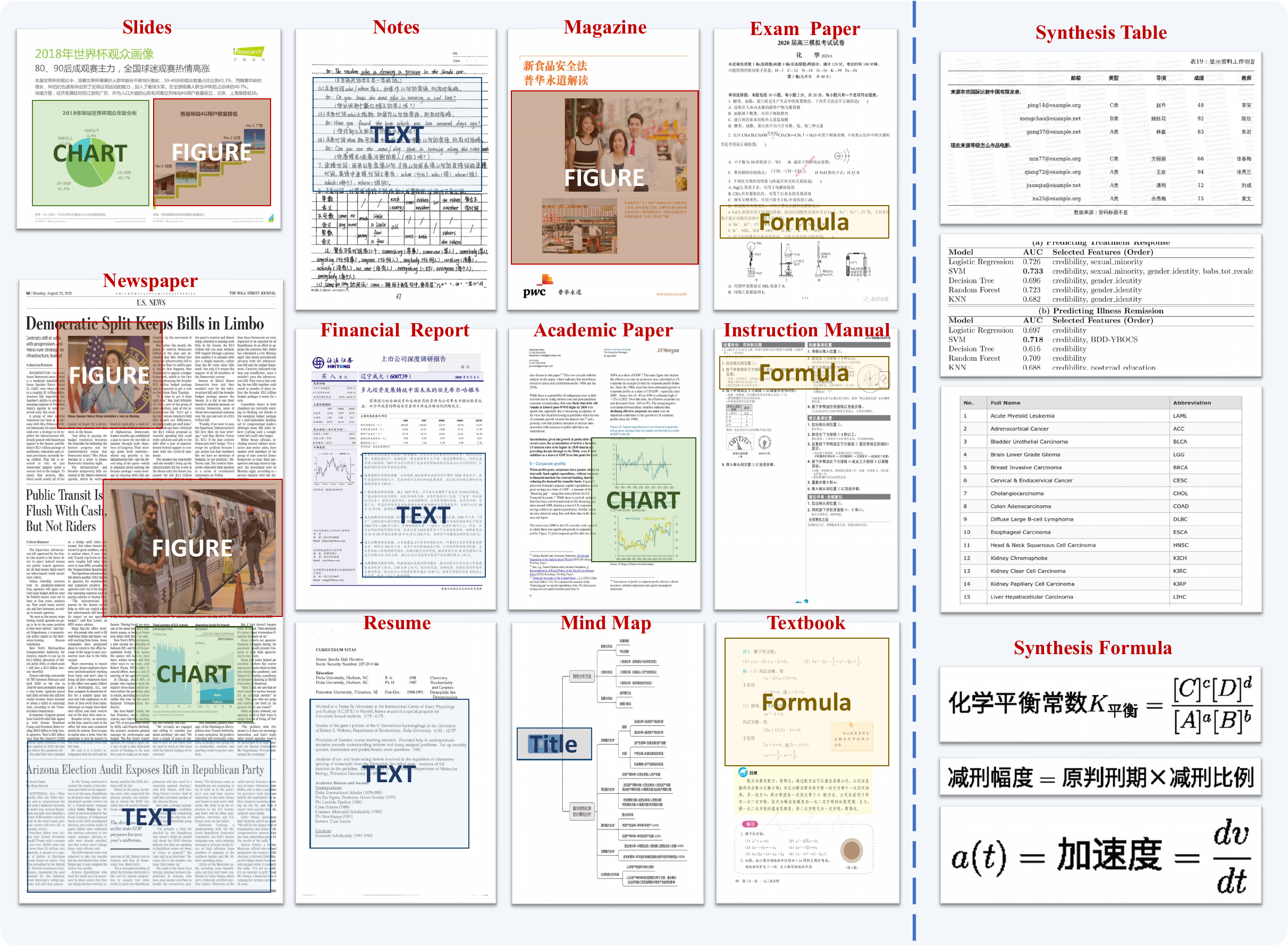}
\caption{Visualization of the MonkeyDoc dataset. MonkeyDoc encompasses more than ten document types and includes our synthesized tables and formulas data.}
\label{fig:doctype}
\end{figure*}

MonkeyDoc is designed to cover a complete range of document parsing tasks, including layout detection, reading order prediction, text recognition, table recognition, formula recognition, and code block recognition. As shown in Fig.~\ref{fig:doctype}, MonkeyDoc spans more than ten diverse document domains and provides high-quality annotations in both Chinese and English, making it a comprehensive and versatile dataset for document parsing. In comparison, previous datasets are often limited to a subset of tasks, a single document type, or monolingual settings. MonkeyDoc uniquely enables multi-task, multi-domain, and bilingual training and evaluation, supporting both fine-grained and holistic document understanding. In the following sections, we provide a detailed description of the data construction process for each stage.

\subsection{Structure Detection}
The task of Structure Detection involves detecting elements such as text, tables, and images within a document and returning their corresponding categories and bounding box coordinates. To construct the Structure Detection dataset, we focus on two main aspects:

\textbf{Filtering from Open-Source Datasets.}
We collect existing open-source Structure Detection datasets, including M\textsuperscript{6}Doc~\cite{m6doc}, DocLayNet~\cite{doclaynet}, D4LA~\cite{d4la}, and CDLA~\cite{cdla}, covering both Chinese and English documents, with a total of 153k pages. We process and filter the original annotations by removing nested bounding boxes and retaining only the largest bounding box along with its category and coordinates. Additionally, we compute the proportion of each bounding box’s area relative to the total page area and remove low-information instances where this ratio is less than 35\%. To address the inconsistency in category annotations across different datasets (e.g., M\textsuperscript{6}Doc~\cite{m6doc} defines 74 categories while D4LA~\cite{d4la} uses 27), we map all category labels to a unified set of classes.

\textbf{Synthesizing Chinese Data.}
 Given the scarcity of Chinese samples in existing open-source datasets, we further collected over 300k pages of Chinese documents spanning financial reports, textbooks, academic papers, and more. These pages were first pre-annotated using an existing Structure Detection model, followed by rule-based post-processing—such as removing nested bounding boxes and filtering low-information instances—and subsequent manual correction and verification to ensure annotation accuracy and consistency. In total, these steps yield 150k high-quality samples.
 
\subsection{Content Recognition}

Content recognition includes text block recognition, table recognition, formula recognition, and code block recognition. To construct data for these tasks, we focus on four key components:

\textbf{Cropping Document Elements.} Leveraging the layouts obtained from Structure Detection, we extracted various types of document elements from the images, yielding a total of 2.5 million cropped samples. Text blocks, formula blocks, and tables were annotated using the advanced model Qwen2.5VL-72B~\cite{qwen2.5-vl} as well as the commercial tool Gemini2.5Pro~\cite{gemini25}.

\textbf{Filtering from Open-Source Datasets.} For table recognition, we filter data from PubTabNet~\cite{pubtabnet} through a series of quality checks, including verifying HTML tag closure, ensuring table structure completeness (e.g., presence of headers), validating the correctness of merged cells, checking header-body alignment, detecting abnormal encoded characters, and performing overall syntax validation. This results in a clean dataset of 470k tables. For formula recognition, we use the UniMER-1M~\cite{unimernet} dataset as the training set, which is constructed from diverse sources, including public datasets such as Pix2tex~\cite{pix2tex}, CROHME~\cite{CROHME2014,CROHME2016,CROHME2019} and HME100K~\cite{hme-100k}, as well as LaTeX expressions collected from Arxiv, Wikipedia and StackExchange. The dataset includes printed, handwritten and screen-captured formulas, ensuring broad coverage of expression styles and complexities. In total, it contains approximately 1.5 million samples.

\textbf{Synthesizing Chinese Data.} Due to the limited availability of Chinese data for table and formula recognition, we manually synthesize additional datasets. For tables, we collect numerous existing table structures featuring diverse multi-row, multi-column, and nested combinations, automatically populate table cells with Chinese characters, and render the results into Chinese table images paired with HTML annotations. For formulas, we synthesize the dataset through two approaches: first, by using large multimodal models to generate commonly used formulas in Chinese across various industries; second, by translating English formulas from the UniMER-1M~\cite{unimernet} dataset into Chinese using large language models and rendering them into Chinese formula images. In total, the synthesized Chinese tables and formulas comprise 135k samples.

\textbf{ArXiv Data Extraction.} To further enrich the dataset, we crawl LaTeX data for tables and formulas from ArXiv. The crawled data is processed using the advanced open-source large language model Qwen2.5-72B~\cite{qwen25} to filter out unnecessary elements such as citations and comments. Additionally, table LaTeX data is converted into HTML using these models and then rendered into images. This ensures the inclusion of high-quality academic data into the dataset, resulting in a total of 411k collected samples.

By combining these four steps, we build a comprehensive and diverse dataset for content recognition, supporting text block, table and formula identification in both Chinese and English.

\subsection{Relation Prediction}
Based on the completed structure detection and content recognition, we further predict the reading order relationships between elements. Existing structure detection models output region-level bounding boxes, which are not directly compatible with line-level reading order prediction models. To bridge this gap, we construct a region-level reading order dataset to train models specifically for this granularity. We collect three types of data for reading order construction: open-source datasets with region level reading order annotations, unlabeled Chinese data with diverse document types, and documents with region level layout annotations, and design dedicated strategies for each.

\textbf{Filtering from Open-Source Datasets.}  
To our knowledge, DocGenome~\cite{docgenome} is the only open-source training dataset that provides region-level reading order annotations. These annotations are generated through an automated labeling process, which may introduce a certain degree of inaccuracy.  To address potential inaccuracies or omissions in the reading order annotations for images and tables, we refine the annotations by strictly associating each image and table with its corresponding caption. We then filter out low-quality data, including pages with unannotated regions or excessive blank areas. After completing the correction and filtering steps, we score samples based on the diversity of element types on each page, incorporating high-scoring pages into our training set. As a result, we obtain 951k samples with high-quality reading order annotations.

\textbf{Manually Annotating Chinese Data.}  
Given the scarcity of Chinese documents in existing datasets, we collected a wide range of Chinese document types, including research reports, academic papers, user manuals, books, test papers, slides, official documents, newspapers and journals, as well as contracts. We manually annotated the region-level reading order of these documents, resulting in 154k samples, to enhance the coverage and diversity of Chinese scenarios in the dataset.

\textbf{Automatic Annotation with Expert Models.}  
Existing layout analysis datasets typically provide region-level bounding box annotations but lack corresponding reading order information. To address this limitation, we leverage the bounding box annotations from these datasets and the line-level LayoutReader~\cite{layoutreader} model to automatically generate region-level reading order labels. Specifically, we first utilize PPOCR~\cite{ppocr} to perform line-wise text recognition within each bounding box, obtaining the positional information of all text lines. Subsequently, LayoutReader predicts the reading order of these text lines. Finally, the region-level reading order is determined by calculating the average predicted order of all text lines within the region. This process produces 228k region-level reading order annotations, effectively enhancing existing layout datasets.

\begin{table*}[ht]
  \centering
\scalebox{1}{
\begin{tabular}{c|ccc}
\midrule
Tab. C& Text & Table & Formula \\ \midrule
1-NED & 99.6 & 99.96 & 99.1    \\ \midrule
\end{tabular}}
  \caption{Comparison between automatically generated annotations and manually corrected ground truth.}    
    \vspace{-5pt}
  \label{tab:human}%
\end{table*}%

\subsection{Comparison with Human Annotation.} 
We implemented a strict quality control procedure for annotation generation. Prior to large-scale generation, we annotated a small subset of samples and manually verified whether the accuracy met our predefined standards. During this pilot phase, we observed that Qwen2.5VL-72B did not achieve satisfactory performance on table recognition. Therefore, we used Qwen2.5VL-72B exclusively for annotating text and formulas, while tables were annotated using Gemini 2.5-Pro and supplemented with automatically synthesized and rendered table data.
After the full dataset was generated, we randomly sampled a substantial portion of instances for manual inspection to identify potential errors. For systematic and recurring errors, we applied rule-based filtering strategies to further refine the annotations. Finally, we randomly selected 100 instances each of text, tables, and formulas for validation. We computed the 1-NED between the original annotations and the manually corrected ground truth, as reported in Tab.~\ref{tab:human}.

\begin{table*}[ht]
\centering
\resizebox{0.9\textwidth}{!}{
\begin{tabular}{c|c|cc|cc|cc|cc|cc|cc}
\toprule
\multirow{2}{*}{\textbf{Vision Encoder}} & \multirow{2}{*}{\textbf{LLM}}
& \multicolumn{2}{c|}{\textbf{Overall}\textsuperscript{\textbf{Edit}}↓} 
& \multicolumn{2}{c|}{\textbf{Text}\textsuperscript{\textbf{Edit}}↓} 
& \multicolumn{2}{c|}{\textbf{Formula}\textsuperscript{\textbf{Edit}}↓} 
& \multicolumn{2}{c|}{\textbf{Table}\textsuperscript{\textbf{TEDS}}↑} 
& \multicolumn{2}{c|}{\textbf{Table}\textsuperscript{\textbf{Edit}}↓} 
& \multicolumn{2}{c}{\textbf{Read Order}\textsuperscript{\textbf{Edit}}↓} \\
\cmidrule{3-14}
 &  & \textit{EN} & \textit{ZH} & \textit{EN} & \textit{ZH} & \textit{EN} & \textit{ZH} & \textit{EN} & \textit{ZH} & \textit{EN} & \textit{ZH} & \textit{EN} & \textit{ZH} \\

\midrule
QwenVit & Qwen2.5-3B & {0.118} & {0.151} & {0.043} & {0.077} &{0.247} & {0.321}  & {84.2} & {89.5} & {0.118} & {0.093} & {0.066} & 0.113 \\
QwenVit & Qwen3-1.7B & 0.131 & 0.149 & 0.044  & 0.074 & 0.293 & 0.326  & 84.7 & 87.9 & 0.125 & 0.087 & 0.062 & 0.110 \\
InternVit &  Qwen3-0.6B & 0.144 & 0.176 & 0.116 & 0.154 & 0.263 & {0.328}  & 83.8 & 87.9 & 0.128 & 0.112 & 0.070 & 0.110 \\

\bottomrule
\end{tabular}}
\caption{The end-to-end evaluation results of different backbones on OmniDocBench.}
\label{tab:multitasks}
\end{table*}

\section{MonkeyOCR with Different Backbones}

We further analyze the performance of MonkeyOCR under different backbone configurations. Specifically, we evaluate multiple encoders (such as QwenViT~\cite{qwen2.5-vl} and InternViT~\cite{internvl}) and LLMs of varying scales (Qwen2.5-3B~\cite{qwen25}, Qwen3-1.7B, and Qwen3-0.6B) trained on MonkeyDoc. Experimental results show that the proposed SRR paradigm and the MonkeyDoc dataset exhibit strong robustness across all backbone settings. Moreover, backbone configurations with larger parameter sizes generally achieve better overall performance.

\section{Comparison on OmniDocBench v1.5}
We further conduct comparisons with other models on OmniDocBench v1.5, which contains 1,355 document images with a more balanced distribution of document and element types. In this updated benchmark, formula recognition is evaluated using the CDM metric. As shown in Tab.~\ref{tab:v1_5}, our model demonstrates strong performance.

\begin{table}[h]
\centering
\begin{tabular}{l|cccccc}
\toprule
\textbf{Model} & \textbf{Overall$\uparrow$} & \textbf{Text$^{\text{Edit}}\downarrow$} & \textbf{Formula$^{\text{CDM}}\uparrow$} & \textbf{Table$^{\text{TEDS}}\uparrow$} & \textbf{Table$^{\text{TEDS-S}}\uparrow$} & \textbf{Read Order$^{\text{Edit}}\downarrow$} \\
\midrule
MonkeyOCR-3B & 91.02 & 0.049 & 89.51 & 88.43 & 92.74 & 0.074 \\
MinerU2.5 & 90.67 & 0.047 & 88.46 & 88.22 & 92.38 & 0.044 \\
dots.ocr & 88.41 & 0.048 & 83.22 & 86.78 & 90.62 & 0.053 \\
Deepseek-OCR & 87.01 & 0.073 & 83.37 & 84.97 & 88.80 & 0.086 \\
\bottomrule
\end{tabular}
\caption{Comprehensive evaluation of document parsing on OmniDocBench v1.5.}
\label{tab:v1_5}
\end{table}

\section{Non-Manhattan Layouts}
We visualize our model’s results on non-rectangular text blocks (Fig.~\ref{fig:nonman} (a)), artistic magazine (Fig.~\ref{fig:nonman} (b)), and paragraph overlapping QR code (Fig.~\ref{fig:nonman} (c)). Non-rectangular text blocks are detected as rectangular regions; in Fig.~\ref{fig:nonman} (b), image order does not affect textual semantics, allowing correct reading order. We do not flag improperly cropped characters, and when structure detection fails, we performs end-to-end recognition on the full page. To avoid overlap in Fig.~\ref{fig:nonman} (c) (5,6,7), the paragraph is split into two blocks.

\begin{figure*}[h]
\centering
\includegraphics[width=1\textwidth]{figs/nmlayout_new.pdf}
\caption{Visualization results on non-Manhattan layouts.}
\label{fig:nonman}
\end{figure*}

\section{More analysis of CPD}
\begin{figure*}[h]
\centering
\includegraphics[width=1\textwidth]{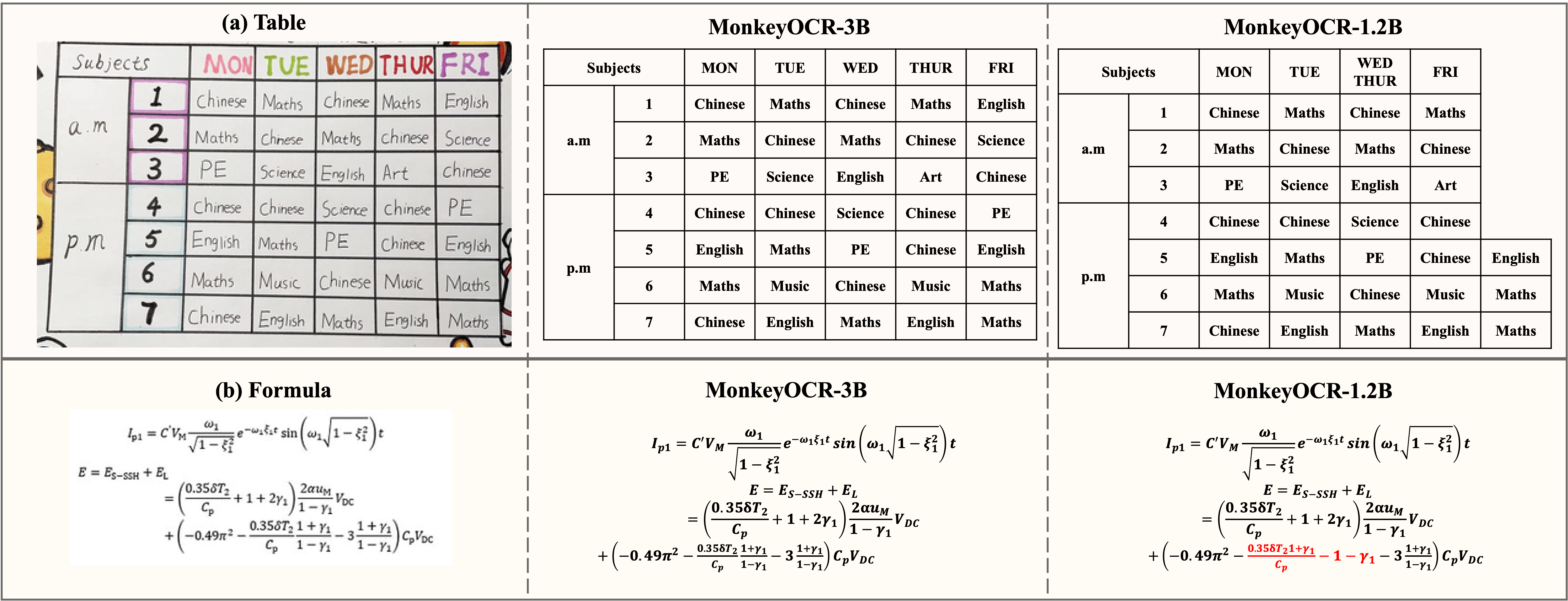}
\caption{Typical failure cases.}
\label{fig:cpd}
\end{figure*}

To further analyze the advantages of CPD, we sampled a subset of the training data to compare CPD with traditional knowledge distillation (KD). Specifically, we initialized the LLM with selected layers from Qwen2.5-3B and used MonkeyOCR-3B as the teacher to train the 1.2B model. Since KD requires additional inference from the 3B model, CPD achieves approximately three times higher training efficiency. Under the same training setup, CPD outperforms KD by 2.17\%. We also examined how the smaller 1.2B model performs across different tasks. For simple tasks, its performance drops only slightly compared to the 3B model, whereas more complex tasks such as formula recognition and table recognition suffer a larger decline. This difference is likely because text recognition only requires content transcription, while tables and formulas involve intricate structures that demand deeper understanding. Visualizing typical failure cases after pruning, we found that errors mainly occur in table and formula structures—for instance, the model misreads table cells (Fig.~\ref{fig:cpd} (a)) and interprets fractions incorrectly as left–right sequences (Fig.~\ref{fig:cpd} (b)).

\section{Limitations}
Pipeline methods (e.g., PPOCR, MinerU) first perform layout detection, followed by text and formula detection, instance-level recognition, merging of recognized elements, and reading-order reconstruction. These approaches rely on at least seven specialized models and complex post-processing. In contrast, our method unifies the process under a triplet paradigm, achieving a significantly better accuracy–efficiency trade-off. Although cumulative errors cannot be fully eliminated, our method substantially mitigates them while remaining efficient.
Unlike pipeline methods that detect individual text instances, our model detects element blocks. Statistics show that inter-block spacing is larger than inter-text spacing (5.2 mm vs. 1.5 mm), making block-level detection more robust. On OmniDocBench v1.5, MinerU produces an average of 97.2 boxes per image, compared to 19.5 for our method.
Furthermore, since reading order in some documents may depend on semantic content, relying solely on spatial and class information can introduce errors. Our method can encode text content into  embeddings and add them with positional embeddings to incorporate semantic cues, which yields an approximate 0.13\% performance improvement on OmniDocBench v1.5, albeit at the cost of increased computational overhead. Nonetheless, our method may still suffer from errors in layout detection, which can propagate to subsequent recognition steps. For example, directly detecting rectangular boxes may be inaccurate for photographed documents, and handling complex layouts that require semantic understanding may remain insufficient. Challenges such as cross-page tables and paragraphs, hierarchical headings, and multi-page document parsing also persist.

\section{Comparison with other Methods}

\begin{figure*}[h]
\centering
\includegraphics[width=1\linewidth]{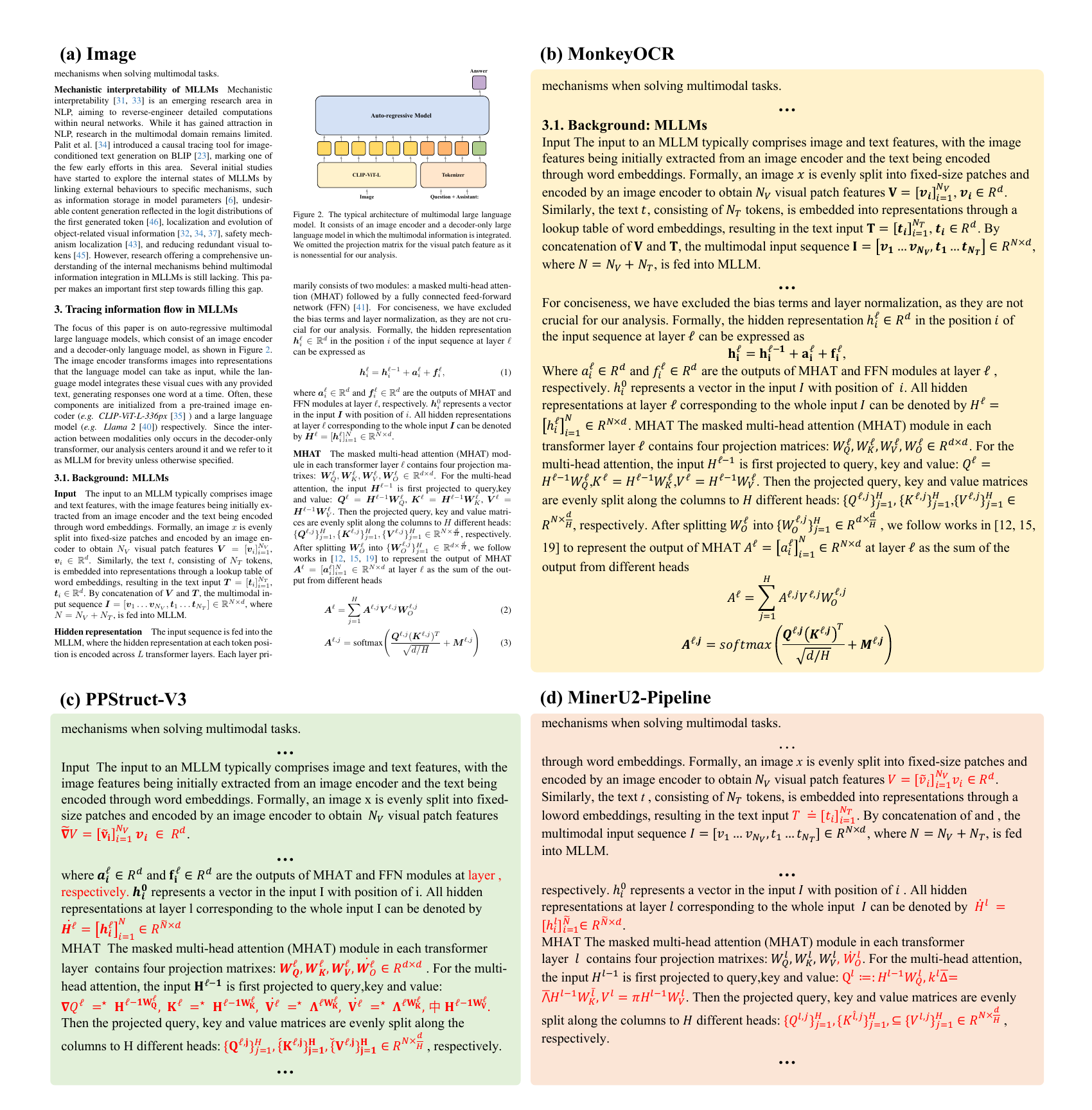}
\caption{Visual comparison with pipeline methods, with errors highlighted in red.}
\label{fig:com1}
\end{figure*}

We further compare our method with the state-of-the-art pipeline approaches, MinerU2-Pipeline~\cite{mineru} and PPStruct-V3~\cite{ppstruct}, with results shown in Fig.~\ref{fig:com1}. The visualizations indicate that these pipeline methods often produce unnecessary superscripts and incorrect symbols when recognizing inline formulas. This issue primarily arises from inaccuracies in inline formula detection, which lead to error accumulation across multiple processing stages. In contrast, our method simplifies the pipeline by directly feeding the text block into an LMM for recognition, resulting in stronger robustness and more accurate outputs.

\begin{figure*}[h]
\centering
\includegraphics[width=1\linewidth]{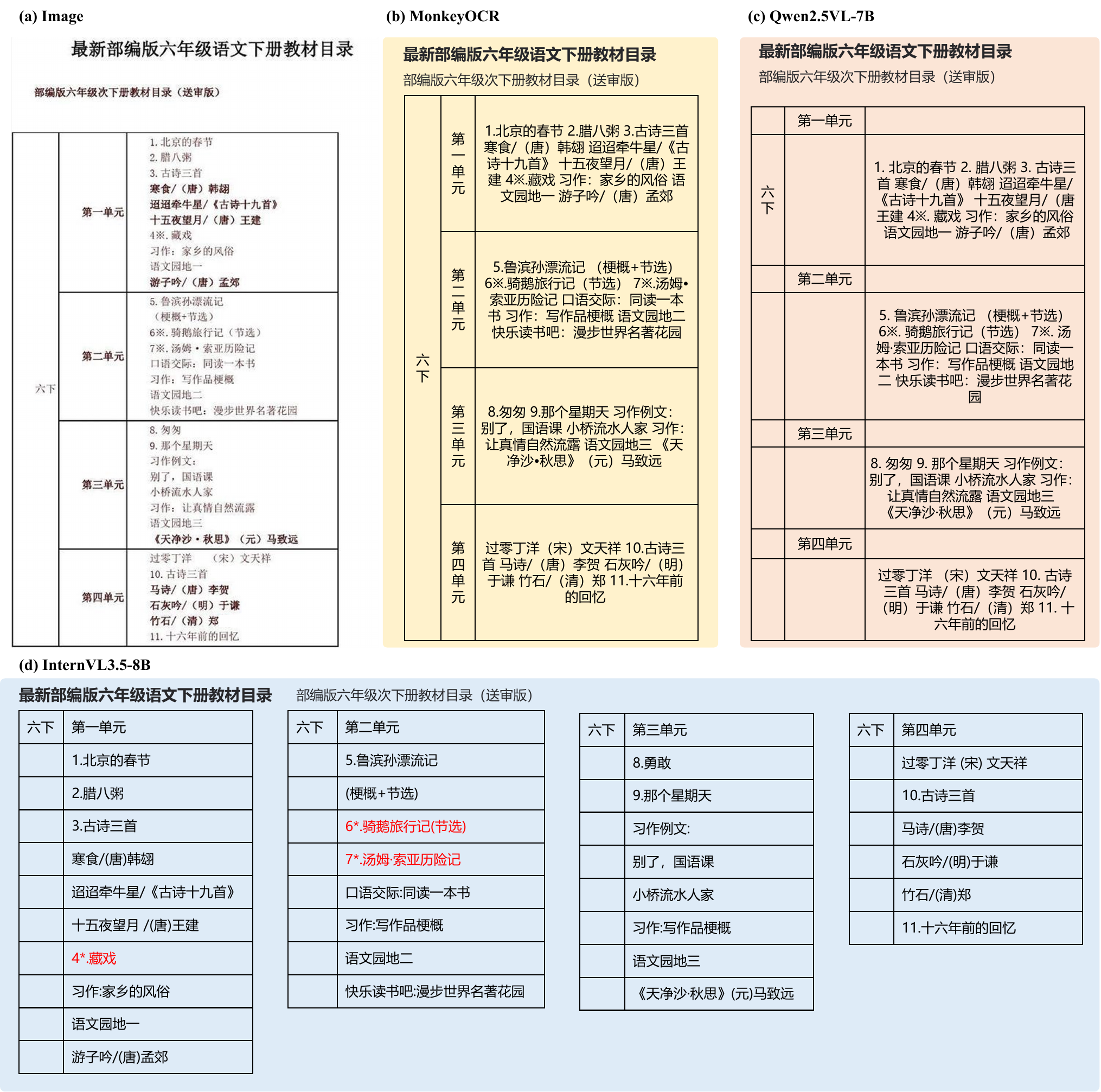}
\caption{Visual comparison with end-to-end models, with errors highlighted in red.}
\label{fig:com2}
\end{figure*}

We compared MonkeyOCR-3B with Qwen2.5VL-7B~\cite{qwen2.5-vl} and InternVL3.5-8B~\cite{internvl}, with results shown in Fig.~\ref{fig:com2}. Despite having fewer parameters, MonkeyOCR-3B accurately reconstructs both the structure and content of the table. In contrast, Qwen2.5VL-7B exhibits errors in table structure recognition, while InternVL3.5-8B not only misidentifies table structures but also makes mistakes in symbol recognition. These results demonstrate that MonkeyOCR-3B excels in table recognition, benefiting from the abundant table recognition samples in the MonkeyDoc dataset.

\end{document}